\documentclass[10pt,twocolumn,letterpaper]{article}

\usepackage{cvpr}      % To produce the REVIEW version
\usepackage{bm}
\usepackage{listings}

\definecolor{cvprblue}{rgb}{0.21,0.49,0.74}
\usepackage[pagebackref,breaklinks,colorlinks,allcolors=cvprblue]{hyperref}
\newcommand{\z}{\bm{z}}
\newcommand{\vect}{\bm{v}}
\def\paperID{*****} % *** Enter the Paper ID here
\def\confName{CVPR}
\def\confYear{2025}

\title{Self-supervised Representation Learning with Local Aggregation for Image-based Profiling}

\author{
	 Siran Dai\textsuperscript{1,2}\hspace{2em} Qianqian Xu\textsuperscript{3,}\thanks{Corresponding authors}\hspace{2em} Peisong Wen\textsuperscript{3,4}\hspace{2em} Yang Liu\textsuperscript{4}\hspace{2em} Qingming Huang\textsuperscript{3,4,*} \\
	{\textsuperscript{1}Institute of Information Engineering, Chinese Academy of Sciences} \\
	{\textsuperscript{2}School of Cyber Security, University of Chinese Academy of Sciences} \\
    {\textsuperscript{3}Institute of Computing Technology, Chinese Academy of Sciences} \\
    {\textsuperscript{4}School of Computer Science and Technology, University of Chinese Academy of Sciences} \\
	{\tt\small daisiran@iie.ac.cn\hspace{2em} \{xuqianqian, wenpeisong20z\}@ict.ac.cn} \\ 
    {\tt\small liuyang232@mails.ucas.ac.cn\hspace{2em} qmhuang@ucas.ac.cn }
}

\begin{document}
\maketitle
\begin{abstract}
Image-based cell profiling aims to create informative representations of cell images. This technique is critical in drug discovery and has greatly advanced with recent improvements in computer vision. Inspired by recent developments in non-contrastive Self-Supervised Learning (SSL), this paper provides an initial exploration into training a generalizable feature extractor for cell images using such methods. However, there are two major challenges: 1) Unlike typical scenarios where each representation is based on a single image, cell profiling often involves multiple input images, making it difficult to effectively fuse all available information; and 2) There is a large difference between the distributions of cell images and natural images, causing the view-generation process in existing SSL methods to fail. To address these issues, we propose a self-supervised framework with local aggregation to improve cross-site consistency of cell representations. We introduce specialized data augmentation and representation post-processing methods tailored to cell images, which effectively address the issues mentioned above and result in a robust feature extractor. With these improvements, the proposed framework won the Cell Line Transferability challenge at CVPR 2025.
\end{abstract}    
\section{Introduction}
\label{sec:intro}
Cell profiling aims to learn meaningful representations of cells that support compound validation in drug discovery and the study of disease mechanisms \cite{Chandrasekaran2020}. Among available methods, image-based profiling from microscopy is the most cost-effective way to produce high-dimensional representations.

Learning representations without human annotations has long been an important goal in computer vision, and recent advances in Self-Supervised Learning (SSL) have brought this closer. Previous research has made significant progress in image-based profiling using computer vision techniques \cite{Caicedo2017, McQuin2018, Bray2016, Sanchez-Fernandez2023, sanchez2022contrastive, Yu2025}, yet obtaining a generalizable feature extractor through self-supervised methods remains an open challenge.

Despite SSL's success on natural images, directly applying standard SSL methods to cell profiling is not suitable. The core issue is the distribution gap between fluorescence microscopy images and natural images. This gap raises two challenges: 1) Natural images typically yield one representation from a single 3-channel input, whereas cell imaging requires one representation from multiple sites in a well and multiple channels, including fluorescent and brightfield images. Leveraging this extra information is crucial. 2) State-of-the-art SSL methods learn view-invariant representations using carefully designed augmentations. Applying these augmentations to cell images is inappropriate and degrades performance due to dimensional collapse \cite{jing2021understanding}.

For the first challenge, since cells within a well belong to the same cell line and receive the same treatment, their images should map to similar points in latent space. Guided by this idea, we propose a new SSL framework with an auxiliary local-aggregation branch that improves cross-site consistency. We then apply feature post-processing to merge outputs from multiple sites within a well into a single representation. The aggregations include multi-granularity merging, interpolation-based merging, cross-site merging, and a cross-plate concentration scheme.

To address the second challenge, we design augmentations tailored to cell images. Because channels correspond to fluorescent dyes, we introduce a channel-aware color jitter. We also simulate imaging noise with a microscope noise augmentation. To improve robustness to cell morphology and anisotropy, we include elastic transformations and random rotations.

We empirically validate our method, named SSLProfiler, by pretraining and evaluating a ViT model on the cell image dataset \cite{chandrasekaran2024three}. We also examine the impact of key components of our proposed method. Our approach achieved first place in the Cell Line Transferability Challenge \cite{cell-line-transferability-challenge-cvdd} at the CVDD workshop, CVPR 2025.

\section{Related Works} \label{sec:related_works} 
\subsection{Self-supervised Learning} 
Initially, SSL aimed at solving pretext tasks designed manually by humans \cite{agrawal2015learning, doersch2015unsupervised, jenni2018self, misra2020self, noroozi2016unsupervised, larsson2017colorization, gidaris2018unsupervised}. Later, contrastive learning methods \cite{moco,cpc,simclr,infonce,mocov2} significantly outperformed supervised pretraining approaches on image-level tasks, making SSL a mainstream method for model pretraining. Recently, non-contrastive learning, which aims to achieve consistency across different views \cite{byol,moco,dino,simclr,simsiam,ayush2021geography}, has attracted increasing interest due to its better ability to generalize. Non-contrastive methods have been successfully applied in various domains, including video representation learning \cite{liu2025future, liu2024not} and medical image processing \cite{asgari2021deep}. Nevertheless, applying SSL to dense prediction tasks \cite{leopart,wen2025semanticconcentrationselfsuperviseddense,dai2025exploringstructuraldegradationdense} or long-tailed scenarios \cite{dai2023drauc} remains challenging. More importantly, how to effectively integrate SSL, especially non-contrastive approaches, into image-based cell profiling remains an open research question.

\subsection{Image-based Cell Profiling} 
Image-based cell profiling has become a powerful method for measuring phenotypic differences across various cellular states. By utilizing high-throughput microscopy and advanced computational methods, this approach extracts detailed morphological features from cell images, providing valuable insights into cellular responses to different conditions. Earlier research in image-based profiling primarily focused on supervised or weakly supervised approaches \cite{McQuin2018,Bray2016,Singh2020,Chandrasekaran2020}. Recent developments have enhanced profiling by integrating contrastive SSL methods \cite{Sanchez-Fernandez2023, sanchez2022contrastive, Yu2025} to obtain richer and more informative representations. Nonetheless, exploring non-contrastive methods in cell profiling may further improves the performance and generalization.

\section{Method}
\label{sec:method}
\subsection{Task Definition}
This paper focuses on extracting meaningful representations from cell images. Specifically, we use images from the \textit{2020\_11\_04\_CPJUMP1} dataset provided in \cite{chandrasekaran2024three}. The dataset includes chemically and genetically perturbed cells from two cell lines: U2OS and A549. Each experiment is performed within a plate that contains multiple wells. Each well represents a distinct experimental condition with unique perturbations. Thus, we define the dataset as $\mathcal D= \{X_w^i\}_{i=1}^N$, where $N$ is the total number of wells. Within each well, images are captured from either 9 or 16 distinct positions, represented as $X_w^i = \{\bm x_j\}_{j=1}^{p_i}$, where $p_i$ is the number of positions in well $i$. Images from each position have 8 channels, consisting of five fluorescent and three brightfield channels.

The goal is to train a robust feature extractor $f$ for each well $X_w$, enabling the learned representation to capture both cellular phenotypic features and causal effects of the applied perturbations. Although SSL methods have been successful with natural images, applying these methods to cell images is challenging for two main reasons: First, cell images provide multiple types of information compared to natural images. For instance, in natural image datasets such as ImageNet \cite{imagenet}, representations are generated based on a single image. In contrast, cell images require integration of data from various positions and multiple channels. Second, a significant distribution gap exists between cell images and natural images. Effective SSL methods heavily rely on data augmentation strategies, such as random cropping and color jittering \cite{simclr}, which must be carefully modified for cell images due to the distinct properties of their channels.

% \subsection{Data Preprocessing}
% The original dataset consists of 16-bit TIFF images. To reduce disk space usage and accelerate data loading during training, we first convert images to 8-bit format, aligning them with natural image standards, using:
% \begin{equation*}
% I(x,y) = \left\lfloor \frac{I(x,y) - \min(I)}{\max(I) - \min(I)} \times 255 \right\rfloor.
% \end{equation*}
% Next, we compute the mean and variance for each channel, denoted by $\bm\mu$ and $\bm \sigma$. During both training and inference, the input images are normalized as follows:
% \begin{equation*}
% \bm x_p = \frac{\bm x_p - \bm\mu}{\bm \sigma}.
% \end{equation*}
\begin{figure*}[!ht]
    \centering
    \includegraphics[width=\linewidth]{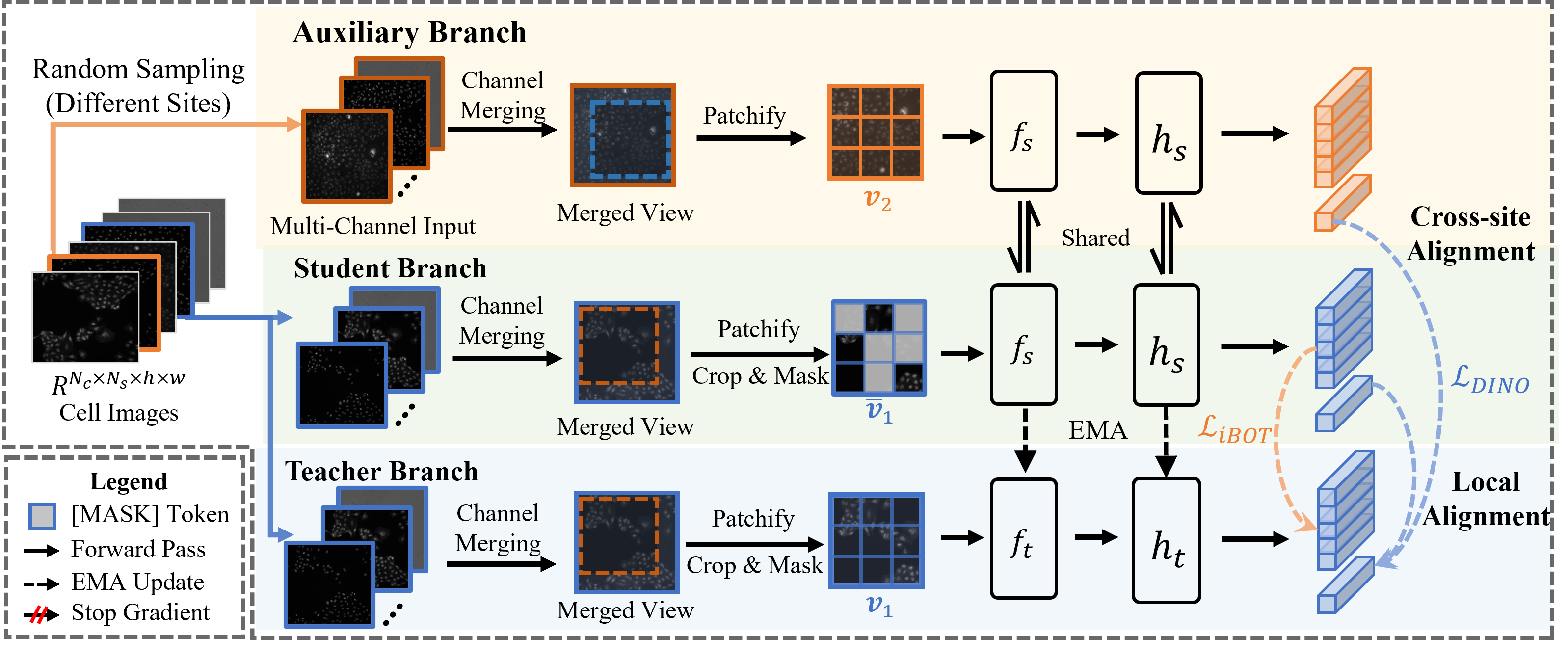}
    \caption{An illustration of the overall framework. }
    \label{fig:overall}
\end{figure*}
\subsection{Self-supervised Learning with Local Aggregation}
While previous SSL frameworks conduct the pretext task mainly by augmenting the input image, they do not fully utilize the properties of cell images. In cell painting, cells within a well are from the same cell line and receive the same treatment. Because these images are phenotypically similar and causally consistent, we argue that they should have similar representations in the latent space, which matches the goal of cell profiling.

Motivated by this idea, we propose an SSL framework that uses local aggregation to improve cross-site consistency. An illustration of the framework is shown in Fig. \ref{fig:overall}.
\subsubsection{Learning Framework}
We start with a classical non-contrastive SSL framework based on a Siamese network, which comprises a student model \( f_s \) and a teacher model \( f_t \). Consistent with standard self-distillation methods \cite{dino, ibot, dinov2}, the student model learns by distilling knowledge from the teacher model, while the teacher model is updated using an Exponential Moving Average (EMA) of the student model parameters. A Vision Transformer (ViT) \cite{dosovitskiyimage} is used as the backbone architecture.

During pretraining, each site in a well is treated as a separate input unit. Unlike natural images, cell images may have up to 8 channels rather than three. Because fluorescent and brightfield channels have very different distributions, and some datasets include only 5 channels, we train two models separately for these channel types and combine their features at test time.

Given the image from a site, we first apply random augmentations to produce two views, $\vect_1$ and $\bar \vect_1$, where $\bar \vect_1$ is additionally masked following \cite{dinov2}. We then introduce an auxiliary branch for local aggregation. We randomly sample an image from another site within the same well and apply random augmentations to obtain the auxiliary view $\vect_2$. Details of the augmentations are provided in the next subsection.

For loss computation, we first employ the DINO v2 losses for intra-image alignment. These include three main components: the DINO loss for instance-level distillation \cite{dino}, the iBOT loss for patch-level reconstruction \cite{ibot}, and the KoLeo loss for regularization \cite{koleo}. Additionally, we enforce local aggregation by pulling the image-level representations of $\vect_1$ and $\vect_2$ closer. The overall loss function is:
\begin{align*}
\mathcal{L}_{total} &= \mathcal{L}_{DINO}(f_s(\bar \vect_1), f_t(\vect_1)) + \mathcal{L}_{DINO}(f_s(\vect_2), f_t(\vect_1)) \\
&+ \lambda_1 \cdot \mathcal{L}_{iBOT}(f_s(\bar \vect_1), f_t(\bar \vect_1)) \\&+ \lambda_2 \cdot \mathcal{L}_{KoLeo}(f_s(\vect_1), f_s(\vect_2)),
\end{align*}
where \(\lambda_1\) and \(\lambda_2\) are hyperparameters. Since the loss functions remain unchanged during training, readers can refer to the original DINO v2 paper \cite{dinov2} for more details.

\subsubsection{Data Augmentation}
\label{sec:dataaug}
The effectiveness of cross-view consistency SSL heavily depends on view generation, specifically data augmentation. Proper data augmentation promotes clustering of similar examples \cite{huang2022towards}, whereas excessively strong augmentations can lead to dimensional collapse \cite{jing2021understanding}. Thus, carefully designing augmentations for cell images is critical.

\noindent\textbf{Adapted Color Jitter}. Considering channels in fluorescent images are more independent than in RGB images, we propose a channel-aware color jitter augmentation that independently adjusts brightness and contrast per channel. Given an input image \(I \in \mathbb{R}^{H \times W \times C}\), the augmented image \(I'\) is computed as follows.

Let \(I_c\) represent the \(c\)-th channel of the input image. For each channel \(c \in \{1, 2, \dots, C\}\), brightness factors \(\beta_c\) and contrast factors \(\gamma_c\) are independently drawn from uniform distributions:
\begin{align*}
\beta_c &\sim \mathcal{U}(\max(0, 1 - \alpha_b), 1 + \alpha_b), \\
\gamma_c &\sim \mathcal{U}(\max(0, 1 - \alpha_c), 1 + \alpha_c),
\end{align*}
where \(\alpha_b\) and \(\alpha_c\) control the maximum brightness and contrast adjustments, respectively.

The augmented image channel \(I_c'\) is first adjusted for brightness:
\begin{align*}
I_c^{(b)} = I_c \cdot \beta_c,
\end{align*}
and subsequently adjusted according to its mean \(\mu_c\):
\begin{align*}
\mu_c &= \frac{1}{HW}\sum_{x=1}^{H}\sum_{y=1}^{W} I_{c}(x,y), \\
I_c' &= (\bm{x}_c^{(b)} - \mu_c) \cdot \gamma_c + \mu_c.
\end{align*}

\noindent\textbf{Additional Augmentations for Cell Images.} To simulate realistic microscopy imaging noise, we apply a microscope noise augmentation, which includes shot noise, dark current, and read noise. These noises are simulated using Poisson and Gaussian distributions. To address cell morphological variability and anisotropy, we also use elastic transformations and random rotations as additional augmentations. A detailed description is presented in the Appendix \ref{sec:app} due to space limitations.

\subsection{Feature Post Processing} \label{sec:post}

So far, we have obtained the feature extractors. However, these extractors produce a representation for each position within a well. In this subsection, we investigate methods to combine these embeddings effectively to obtain strong representations for each well.

\subsubsection{Multi-Granularity Information Merging} \label{sec:multi-granularity}

Since we use ViT models as the backbone, they naturally provide representations at different levels: the image-level ([CLS] token) and the dense-level (patch tokens). To effectively use these multi-level features, we concatenate the [CLS] token with the average of the patch tokens, forming the final output for position $p$: \begin{equation*} \z_p = \text{concat}(\z_p^{\text{CLS}}, \z_p^{\text{Patch}}). \end{equation*}

\subsubsection{Multi-Position Information Merging} \label{sec:cross_plate}

Next, we combine representations from multiple positions into a single well-level representation. In the \textit{2020\_11\_04\_CPJUMP1} dataset, images are captured at either 9 or 16 positions. Typically, these representations can be merged through averaging or concatenation. However, direct concatenation is not suitable for cell profiling tasks because the representations from different wells would have varying dimensions, complicating downstream tasks. To solve this issue, we assume that positions are sampled in a symmetrical pattern. Based on this assumption, wells with 16 images are first reshaped into a $4 \times 4$ matrix, then interpolated into a $3 \times 3$ matrix. After this adjustment, all wells contain the same number of representations (9 representations), allowing for effective concatenation. Practically, we observe that concatenation typically provides better results compared to averaging.

\subsubsection{Cross-Plate Representation Alignment}

To further enhance the consistency of learned representations across different experiments, we apply a simple yet effective method. Specifically, we observed in our dataset that perturbations are linked to specific well positions, meaning that the same well positions across different plates are affected by identical compounds. Thus, aligning representations from the same well positions improves the causal relationship between learned representations and compound-induced perturbations. Considering representations from each well position as clusters, we first calculate the cluster centers: \begin{equation*} \bm{\mu}_{w} = \frac{1}{N_p}\sum_{j=1}^{N_p} \z_{j}^w, \end{equation*} where $N_p$ is the number of plates, and $\z_j^w$ denotes the well representation at position $w$ on plate $j$. Then, we shift each well representation toward its cluster center as follows: \begin{equation*} \z_{j}^w = \alpha \cdot \z_{j}^w + (1 - \alpha ) \cdot \bm{\mu}_{w}, \end{equation*} where $\alpha$ is a hyperparameter controlling the degree of alignment. We find that this strategy enhances the causal relationship between cell representations and compounds, particularly excelling in the CVDD challenge \cite{cell-line-transferability-challenge-cvdd}. However, this may slightly decrease the generalization ability by reducing the phenotypic information captured in cell representations.
\section{Experiments}
\label{sec:exp}
\subsection{Experiment Setup}
\textbf{Model Pretraining}. We pretrained a ViT-Small-14 model from scratch for 100 epochs on the \textit{2020\_11\_04\_CPJUMP1} batch from \cite{chandrasekaran2024three}. For the feed-forward network, we employed SwiGLU \cite{shazeer2020glu}. We set the batch size to 128 and used a learning rate of $2\cdot 10^{-5}$, with a warmup period of 10 epochs. We trained two separate models for fluorescent and brightfield channels, respectively, and combined their outputs using the post-processing method described in Sec \ref{sec:post}.

\noindent\textbf{Evaluation Protocol}. We adopted the evaluation setup from the Cell Line Transferability challenge at CVPR 2025 \cite{cell-line-transferability-challenge-cvdd}, which uses a $k$-NN classifier as the downstream task. Specifically, we trained a $k$-NN classifier on the learned representations of wells to associate these representations with compound perturbations. We applied $K$-fold cross-validation to split the dataset into training and evaluation subsets. To assess the robustness of the representations, the evaluation included tests both within and across different cell lines. Further details can be found in \cite{cell-line-transferability-challenge-cvdd}.
\subsection{Main Results}
We present an analysis of the key components in Tab. \ref{tab:improvements}. In this table, `Local' refers to our reimplemented evaluation, while `leader board' indicates the official results from the challenge. The baseline submission refers to example representations provided by the challenge organizers. Below, we detail each component included in our analysis: 
\begin{itemize} 
    \item \textbf{8 channels input} indicates the adaptation of the DINO v2 framework to the cell image dataset, with only the number of channels changed. 
    \item \textbf{Patch representations} refers to the methods in Sec \ref{sec:multi-granularity}. 
    \item \textbf{Separate training} means we trained two separate models for fluorescent and brightfield channels. This approach improved both training stability and model performance. 
    \item \textbf{Training res 518} means we increased the training resolution of the global view to $518 \times 518$. This significantly improved the results since higher resolution images are necessary to capture detailed information about small cells. 
    \item \textbf{Adapted augmentations} refers to the augmentation techniques described in Sec \ref{sec:dataaug}, confirming the effectiveness of our approach. 
    \item \textbf{Local aggregation} refers to the local aggregation technique described in Sec \ref{sec:method}.
    \item \textbf{ViT-Base-14} means we increased the backbone model size, achieving better performance. 
    
  \end{itemize}
    
\definecolor{lightred}{RGB}{255,150,150}
\definecolor{lightgreen}{RGB}{150,255,150}

\newcommand{\uparrowcolor}[1]{\textcolor{lightred}{$\uparrow$} \textcolor{lightred}{#1}}
\newcommand{\downarrowcolor}[1]{\textcolor{lightgreen}{$\downarrow$} \textcolor{lightgreen}{#1}}

\begin{table}[t]
  \centering
  \caption{Ablation study of the key differences between DINO v2 and SSLProfiler.}
  \label{tab:improvements}
  \begin{tabular}{lll}
    \toprule
     & \multicolumn{1}{l}{Local} & \multicolumn{1}{l}{Leader board} \\
    \midrule
    Baseline submission & & 24.1 \\
    DINO v2 & - & - \\
    ~~+~8 channels input &20.3 & 19.5 \\
    ~~+~Patch representations & 20.9 \uparrowcolor{0.6}& 20.5 \uparrowcolor{1.0} \\
    ~~+~Separate training & 24.9 \uparrowcolor{4.0} & 23.4 \uparrowcolor{2.9} \\
    ~~+~Training res 518 & 34.2 \uparrowcolor{9.3}& 31.1 \uparrowcolor{7.7} \\
    ~~+~Adapted augmentations & 35.4 \uparrowcolor{1.2}& - \\
    ~~+~Local aggregation & 38.6 \uparrowcolor{3.2}& - \\
    ~~+~ViT-Base-14 & 40.0 \uparrowcolor{1.6}& OOM \\
   
    \bottomrule
  \end{tabular}
\end{table}

\section{Conclusion}
In this paper, we presented an initial attempt to utilize the success of SSL to enhance cell profiling capabilities. However, the considerable differences between natural images and cell images make this task challenging. Using DINO v2 as our starting point, we investigated several important factors, including data preprocessing, model pretraining, and representation post-processing. Our experiments confirmed the effectiveness of each component. In future work, we plan to further explore ways to improve training algorithms for cell profiling.
{
    \small
    \bibliographystyle{ieeenat_fullname}
    \bibliography{main}
}

\clearpage
\setcounter{page}{1}
\maketitlesupplementary
\section{Details of Additional Augmentations for Cell Images.}
\label{sec:app}
\subsection{Microscope Noise Data Augmentation}

The microscope noise augmentation introduces realistic imaging noise components into the original microscopy image, modeled mathematically as follows:

Given an input image \( I(x,y) \) with intensity values normalized in the range \([0, 1]\), the augmented noisy image \( I_{\text{noisy}}(x,y) \) is defined by incorporating three types of noise: shot noise, dark current, and read noise.

Specifically, the augmented image is computed as:
\begin{align*}
    &I_{\text{photon}}(x,y) = \frac{I(x,y)}{\sigma_{\text{shot}}}, \\
    &\hat{I}_{\text{photon}}(x,y) \sim \text{Poisson}\left(I_{\text{photon}}(x,y)\right), \\
    &I_{\text{noisy}}(x,y) = \hat{I}_{\text{photon}}(x,y) \cdot \sigma_{\text{shot}} + \sigma_{\text{dark}} + \mathcal{N}(0, \sigma_{\text{read}}^2), \\
    &I_{\text{noisy}}(x,y) = \text{clip}\left(I_{\text{noisy}}(x,y), 0, 1\right),
\end{align*}
where \( \mathcal{N}(0, \sigma_{\text{read}}^2) \) represents Gaussian noise with zero-mean and variance \(\sigma_{\text{read}}^2\).

\noindent\textbf{Hyper-parameters:}  
\begin{itemize}[nosep,leftmargin=1.5em]
    \item \( \sigma_{\text{shot}} \) denotes the shot noise scaling factor (default 0.1).
    \item \( \sigma_{\text{dark}} \) represents the dark current noise level (default 0.05)
    \item \( \sigma_{\text{read}} \) denotes the standard deviation of the Gaussian read noise (default 0.01).
\end{itemize}

This augmentation simulates realistic microscopy imaging conditions, providing robustness against various noise artifacts commonly encountered in practical scenarios.

\subsection{Elastic Deformation}
We augment each training image $I\in\mathbb{R}^{H\times W\times C}$ with probability $p$ by an elastic transformation.  
First, two independent displacement fields
\[
\delta_x,\;\delta_y \;\; \sim\; \alpha\,\mathcal{G}_{\sigma}\!\ast\!\mathcal{N}(0,1)^{H\times W},
\]
are generated by convolving zero-mean, unit-variance Gaussian noise with a Gaussian kernel of standard deviation $\sigma$ and subsequently scaling by the elasticity coefficient $\alpha$.  Here $\mathcal{G}_{\sigma}$ denotes the Gaussian smoothing operator and “$\ast$’’ is the convolution. A dense mesh grid of base coordinates is constructed as:
\[
\mathbf{x}(u,v) = (v,\,u),~~ 
(u,v)\in\{0,\dots,H-1\}\times\{0,\dots,W-1\},
\]
and perturbed to obtain the deformed sampling locations
\[
\tilde{\mathbf{x}}(u,v)=\mathbf{x}(u,v)+\bigl(\delta_x(u,v),\,\delta_y(u,v)\bigr).
\]
The augmented image $\tilde{I}$ is obtained by applying channel-wise bilinear interpolation:
\[
\tilde{I}_c(u,v)=I_c\bigl(\tilde{\mathbf{x}}(u,v)\bigr), 
\quad c=1,\dots,C .
\]

\noindent\textbf{Hyper-parameters:}  
\begin{itemize}[nosep,leftmargin=1.5em]
  \item $\alpha$: magnitude of deformation (default $1200$).  
  \item $\sigma$: spatial smoothness of the displacement field in pixels (default $40$).  
  \item $p$: probability of applying ED to a sample (default $0.5$).  
\end{itemize}

The operation preserves image topology while introducing smooth, non-linear distortions, effectively enlarging the training distribution by simulating natural variations of object shape.

\end{document}